\documentclass[letterpaper, 10 pt, conference]{ieeeconf}  % Comment this line out if you need a4paper

\IEEEoverridecommandlockouts                              % This command is only needed if
\usepackage{graphics} % for pdf, bitmapped graphics files
\usepackage{epsfig} % for postscript graphics files
\usepackage{times} % assumes new font selection scheme installed
\usepackage{amsmath} % assumes amsmath package installed
\usepackage{amssymb}  % assumes amsmath package installed

\usepackage{algorithm}
\usepackage{algorithmic}

\usepackage{subfigure}

\usepackage{multirow}

\usepackage{booktabs}

\title{\LARGE \bf
Hindsight Generative Adversarial Imitation Learning }

\author{Naijun Liu$^{1,2}$, Tao Lu$^{1,2}$,  Yinghao Cai$^{1,2}$, Boyao Li$^{1,2}$, and Shuo Wang$^{1,2,3}$
\thanks{----------------------------------- }
\thanks{ Preprint. Work in progress.}% <-this % stops a space
\thanks{  $^1$State Key Laboratory of Management and Control for Complex Systems,
        Institute of Automation, Chinese Academy of Sciences  $^2$University of Chinese Academy of Sciences $^3$Center for Excellence in Brain Science and Intelligence Techology of the
Chinese Academy of Sciences.}
}

\begin{document}
\maketitle
\thispagestyle{empty}
\pagestyle{empty}

\begin{abstract}  %\footnote{}
Compared to reinforcement learning, imitation learning (IL) is a powerful paradigm for training agents to learn control policies efficiently from expert demonstrations. However, in most cases, obtaining demonstration data is costly and laborious, which poses a significant challenge in some scenarios. A promising alternative is to train agent learning skills via imitation learning without expert demonstrations, which, to some extent, would extremely expand imitation learning areas. To achieve such expectation, in this paper, we propose Hindsight Generative Adversarial Imitation Learning (HGAIL) algorithm, with the aim of achieving imitation learning satisfying no need of demonstrations. Combining hindsight idea with the generative adversarial imitation learning (GAIL) framework, we realize implementing imitation learning successfully in cases of expert demonstration data are not available. Experiments show that the proposed method can train policies showing comparable performance to current imitation learning methods. Further more, HGAIL essentially endows curriculum learning mechanism which is critical for learning policies.

\end{abstract}

%%%%%%%%%%%%%%%%%%%%%%%%%%%%%%%%%%%%%%%%%%%%%%%%%%%%%%%%%%%%%%%%%%%%%%%%%%%%%%%%
\section{INTRODUCTION}

Reinforcement learning (RL) has appeared as a promising method for solving complex decision-making tasks, such as video games\cite{c1}, robot manipulation \cite{c2}\cite{c3}, and autonomous driving \cite{c4}. However, devising appropriate reward functions can be quite challenging for many applications \cite{c4n}. Inverse reinforcement learning (IRL) \cite{c5} addresses the problem of learning reward functions from demonstration data, which is often considered as a branch of imitation learning (IL) \cite{c6}. Instead of learning reward functions, other methods of imitation learning were proposed to learn a policy directly from expert demonstrations.

Prior works addressed the IL problems by behavior cloning (BC) which reduces learning a policy from expert demonstrations to supervised learning \cite{c7}. However, covariate shift always gives rise to compounding errors \cite{c8}. To overcome the drawbacks of BC, generative adversarial imitation learning (GAIL) algorithm \cite{c9} was proposed based on the formulation of generative adversarial networks (GAN) \cite{c10}, where the generator is trained to generate expert-like samples and the discriminator is trained to distinguish between generated and real expert samples. GAIL is an appealing approach which is a highly effective and efficient learning framework for policy learning with unknown reward.
\begin{figure}[t] %[thpb]
\setlength{\abovecaptionskip}{0pt}
\setlength{\belowcaptionskip}{0pt}
\centering
\includegraphics[scale=0.28]{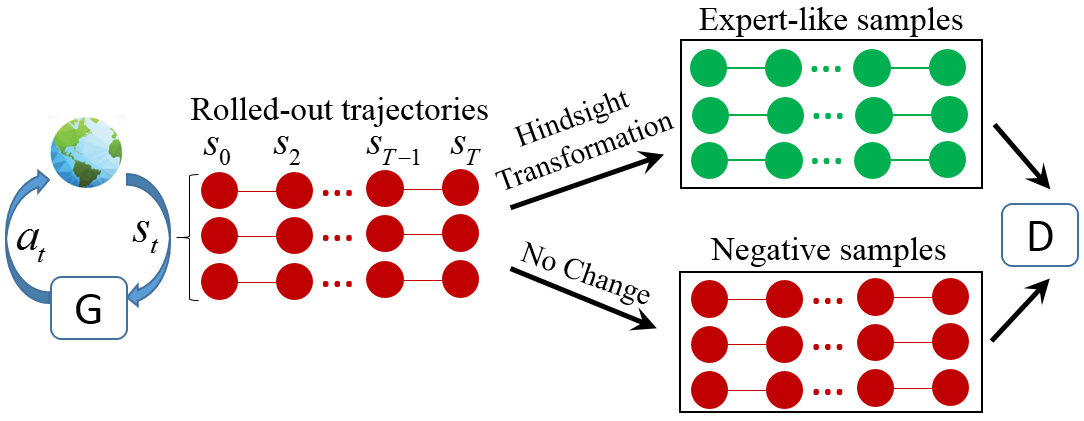}
\caption{The illustration of hindsight generative adversarial imitation learning (HGAIL) algorithm. Rolled-out trajectories are produced via policy (generator) G interacting with the environment. Expert-like demonstration data is converted from rolled-out trajectories with hindsight transformation technique. Regarding rolled-out trajectories as negative samples and treating expert-like demonstration data as positive samples, the discriminator is trained to distinguish between expert-like samples and negative samples. }
\label{fig1}
\end{figure}

Inevitably, demonstration data or usually high-quality demonstration data have to be provided for  imitation learning paradigm \cite{c11}. However, gathering enough high-quality expert demonstrations in many scenarios is usually costly and difficult. To this end, some methods were proposed to apply imitation learning algorithms training policies by leveraging demonstration data as few as possible even reducing to a single demonstration\cite{c12}.

Taking this step further and pursuing an alternative learning paradigm, in this paper, we consider the problem of whether the imitation learning algorithm can be employed successfully without demonstration available, and the final learned policy can show comparable performance to current imitation learning algorithms, which is in high promising.

In order to solve the suggested problem, a feasible way is to make the proposed algorithm intelligently self-synthesize expert-like demonstration data. To do so, we propose hindsight generative adversarial imitation learning (HGAIL) algorithm, which combines the idea of hindsight inspired from psychology \cite{c13} and hindsight experience replay (HER) \cite{c14}  with GAIL into a unified learning framework.
In the process of adversarial training, as illustrated in Figure 1, rolled-out trajectories are generated by policy G interacting with the environment. Expert-like samples are converted from rolled-out trajectories based on hindsight transformation, where the rolled-out trajectories are directly treated as negative samples without any change, which satisfies the requirements for training the discriminator and the generator. Our experimental results show that the  proposed HGAIL allows the agent training to proceed on the rails with no demonstration data provided. The final learned policy shows comparable performance compared with current imitation learning methods.

Furthermore, our HGAIL algorithm essentially endows curriculum learning mechanism in the adversarial learning procedure. At different optimizing iteration steps, expert-like demonstration data are synthesized from different levels rolled-out trajectories data. Therefore, the rolled-out trajectories data and self-synthesized expert-like data are always appropriate for adversarial training, which make the process of adversarial policy learning stable and efficient. As shown in our experiments, the curriculum mechanism is crucial in improving the performance of the final learned policy.

In summary, our main contribution is a method of achieving imitation learning with no demonstration data available. Expert-like demonstration data are self-synthesized with the hindsight transformation mechanism under the proposed HGAIL learning framework. In addition, our method dynamically transforms the rolled-out trajectories data into expert-like data in the training process which ensures the hindsight transformed data is at the appropriate level for adversarial policy learning. To some extent, this latent learning mechanism automatically forms curriculum learning which is of great benefits for improving the performance of the learned policy. Our proposed HGAIL algorithm is also sample-efficiency as we only need rolled-out trajectories data generated by the agent interacting with the environment. No demonstration data or any other external data are required.

%%%%%%%%%%%%%%%%%%%%%%%%%%%%%%%%%%%%%%%%%%%%%%%%%%%%%%%%%%%%%%%%%%%%%%%%%%%%%%%%
\section{RELATED WORK}
\subsection{ Imitation Learning}
Imitation learning algorithms can be classified into three broad categories: behavior cloning (BC), inverse reinforcement learning (IRL), and generative adversarial imitation learning (GAIL).

Behavior cloning reduces the imitation learning to supervised learning, which is simple and easy to be implemented \cite{c7}. However, BC needs huge amount of high-quality expert demonstrations \cite{c11}.

Inverse reinforcement learning addresses the imitation learning problem by inferring a reward function from demonstration data and then using the learned reward function to train policy. Prior works in IRL includes maximum-margin \cite{c15}\cite{c16} and maximum-entropy \cite{c17}\cite{c18}\cite{c19}formulations. 

Generative adversarial imitation learning (GAIL) \cite{c9} is a recent imitation learning method inspired by generative adversarial networks (GAN) \cite{c10}. Another similar frameworks called guided cost learning (GCL) have also been proposed \cite{c20} for inverse reinforcement learning. As training GAIL is notoriously unstable, lots of works focus on improving stability and robustness by learning semantic policy embeddings \cite{c21} , via kernel mean embedding \cite{c22}, or enforcing information bottleneck to constrain information flow in the discriminator \cite{c23}. More recent works extends the learning framework by improving on learning robust reward with state only \cite{c24} or with state-action pairs \cite{c25} in transferred setting for new policy learning.

Other works deal with actions being not available in the demonstration data \cite{c26}, or capturing the latent structure underlying expert demonstrations \cite{c27} for imitation learning.

\subsection{Hindsight Experience Replay}
Hindsight experience replay (HER) \cite{c14} is proposed for dealing with sparse rewards in reinforcement learning. The key insight of Hindsight Experience Replay (HER) is that even in failed rollouts where no valuable reward was obtained, the agent can transform them into successful ones by assuming that a state it saw in the rollout was the actual goal.
Recent works have improved the performance of HER by rewarding hindsight experiences more \cite{c28} , combining curiosity and prioritization mechanism \cite{c29}, or calculating trajectories energy based on work-energy in physics \cite{c30}. An extension of HER called dynamic hindsight experience replay (DHER) \cite{c31} is proposed to deal with dynamics goals.

\subsection{Learning with Few Data}
Generally, training policies with imitation learning method needs expert demonstration data even with huge number or high quality. In some training scenarios, obtaining expert demonstration is not an easy thing \cite{c11}. Lots of work has been emerged to achieve the goal of making imitation learning algorithms work well with fewer demonstration data \cite{c32}. Such as meta-learning framework \cite{c33}\cite{c34}\cite{c35}, neural task programming \cite{c37}, and combining reinforcement with imitation learning \cite{c39}. Zero-shot learning is proposed for addressing visual demonstration without action \cite{c40}.

In recent works, self-imitation learning methods \cite{c41}\cite{c42} are proposed to train policies to reproduce the agent¡¯s past good experience without external demonstration provided. \cite{c43} proposes Generative Adversarial Self-imitation Learning (GASIL) method, which encourages the agent to imitate past good trajectories via generative adversarial imitation learning framework.

Instead of choosing the top-K trajectories according to episode return as the positive samples in GASIL \cite{c43}, we employs the hindsight idea to directly transform the generator data to expert-like demonstration data. Experiment shows that our proposed method outperforms GASIL in robot's reaching and grasping target objects scenarios (section IV-A).
%%%%%%%%%%%%%%%%%%%%%%%%%%%%%%%%%%%%%%%%%%%%%%%%%%%%%%%%%%%%%%%%%%%%%%%%%%%%%%%%
\section{METHOD}
To outline our method, we first consider a standard GAIL learning framework consisting of a policy (generator) $\pi_\theta$ , and discriminator ${D_\omega }$, parameterized with $\theta$ and $\omega$ respectively. The goal of the policy is to generate rolled-out trajectories similar to demonstration trajectories, and the discriminator is to distinguish between state-action pairs sampled from the expert demonstration trajectories and the generator trajectories. The generator and the discriminator are optimized with the following objective function
\begin{equation}
\begin{split}
\mathop {\min }\limits_\theta  \mathop {\max }\limits_\omega L(\theta ,\omega )=
& {{\rm E}_{{\pi _\theta }}}[log({D_\omega }(s,a))] + \\
& {{\rm E}_{{\pi _E}}}[log(1 - {D_\omega }(s,a))] - \lambda H({\pi _\theta })
\end{split}
\end{equation}
where ${\pi _E}$ is expert policy, $H({\pi _\theta })$ is the casual entropy of policy ${\pi _\theta }$ which encourage the policy sufficiently to explore the action space, and  $\lambda $ is the regularization weight.

In the concrete implementation of the proposed HGAIL approach, policy and discriminator are represented by multi-layer neural networks. The output ${\pi_\theta}( \cdot)$ of the policy network parameterizes the Gaussian distribution policy $N({\pi_\theta}(\cdot), \delta)$, where ${\pi_\theta}( \cdot)$ is the mean and $\delta$  is the covariance. At the beginning of each episode, our agent sample a goal $g\in G$ and an initial state ${s_0}\in S$. At time step $t$, the agent take an actions  $a_t$ sampled from the Gaussian policy ${a_t}\sim N({\pi _\theta }({s_t}||g), \delta )$ based on current policy ${\pi _\theta }$, state ${s_t}$ and goal $g$, where $||$ denotes concatenation. Then the agent moves to next state ${s_{t + 1}}$ based on transition dynamics $p({s_{t+1}}/{s_t},{a_t})$, and receives reward ${r_{dis}({s_t},{a_t})}$ given by the discriminator. At the end of each episode, a trajectory sequence ${\tau ^i} \leftarrow < {s_{\rm{0}}}{\rm{||}}g{\rm{,}}{a_{\rm{0}}}{\rm{,}}{s_{\rm{1}}}{\rm{||}}g{\rm{, }}{a_{\rm{1}}}{\rm{,}} \cdots {\rm{, }}{s_T}{\rm{|| }}g > $ is generated, where $T$ is the length of trajectory. Repeating above procedure $N$ times, we obtain rolled-out trajectories $\tau\leftarrow\{{\tau ^0},{\tau ^1},\cdots,{\tau^N}\}$.

In order to train the discriminator without expert demonstrations, our method leveraging the hindsight transformation technique (as shown in Algorithm 1) to convert the rolled-out trajectories $\tau$ into expert-like trajectories $\tau_h$.

\renewcommand{\algorithmicrequire}{ \textbf{Input:}} %Use Input in the format of Algorithm
\renewcommand{\algorithmicensure}{ \textbf{Output:}} %UseOutput in the format of Algorithm

\begin{algorithm}
\caption{Synthesizing expert-like demonstrations with hindsight transformation}
\begin{algorithmic}[1]
\REQUIRE
Rolled-our trajectories $\tau \rightarrow \left\lbrace \tau_0, \tau_2, \cdots, \tau_N\right\rbrace$ \\
Hindsight transformation probability  $p_{ht}$ \\
Time index set $H_I$ for hindsight transformation
\ENSURE
Hindsight transformed trajectores  $\tau_{h}$

\FOR{each trajectory $\tau^i$ in  $\tau$ }
\STATE{Set $H_I=\emptyset$}
\FOR{each time step $t$ in $\tau^i$ } 		
\STATE{ Append $t$ to  $H_I$ with probability $p_{ht}$ }
\ENDFOR
\FOR{$j$ in $H_I$}
\STATE{Randomly sample a achieved positon $p_{j}$ from $s_{j}$ to $s_T$}
\STATE{Set the new goal of state $s_j$ to be $p_{j}$ }
%\STATE{obtaining hindsight transitioned step set $t^{h}_i \leftarrow \left\lbrace t_0^{h},t_1^{h}, \cdots, t_k^{h}\right\rbrace $ , $k\leq T$}

\ENDFOR
\STATE{hindsight transformed trajectory $\tau^i_{h}$}
\ENDFOR

\STATE{ $\tau_{h} \leftarrow \left\lbrace \tau^0_{h}, \tau^1_{h}, \cdots, \tau^N_{h}\right\rbrace $}
\RETURN $\tau_{h}$
\end{algorithmic}
\end{algorithm}

More specifically, the detail steps for self-synthesizing expert-like trajectories $\tau_{h}$ from rolled-out trajectories $\tau$ can be described as follows: firstly, for each trajectory ${\tau^i}$ in $\tau$, we choose each time step $t$ with probability $p_{ht}$ for making hindsight transformation, where $t\subseteq[0,T]$. All the chosen time steps in ${\tau^i}$ for hindsight transformation is appended to ${H_I}$. Secondly, for every time step $j$ in ${H_I}$, we randomly set the new goal of state ${s_j}$ with the achieved position ${p_l}$  at state ${s_l}$, where  $l$ is randomly chosen from time step $j$ to $T$ in ${\tau^i}$. In other words, we randomly set new goal of state ${s_j}$ with the position achieved after observe state ${s_j}$. Then, we succeed in transforming the rolled trajectory ${\tau^i}$ into expert-like trajectory ${\tau^i_{h}}$. Repeating above procedure until all trajectories are transformed.

The policy (generator) ${\pi_\theta}$ is optimized with policy gradient method proximal policy optimization PPO [44]. The objective function is
\begin{equation}\label{2}
  \mathop {\min }\limits_\theta  {{\rm E}_\tau }[log{D_\omega }(s,a)] - {\lambda _1}H({\pi _\theta })
\end{equation}
The gradient is given by
\begin{equation}\label{3}
{\nabla _\theta }L = {{\rm E}_\tau }[{\nabla _\theta }log{\pi _\theta }(a|s)Q(s,a))] - {\lambda _1}H({\pi _\theta })
\end{equation}
where $Q(s,a) = {E_\tau }[\log {r_{dis}}(s,a)|{s_0} = s,{a_0} = a]$
is action-value function, ${r_{dis}}(s,a)$ is the reward function output from the discriminator.

The discriminator is optimized with the following function via minimizing the cross entropy
\begin{equation}\label{4}
 \mathop {\max }\limits_\omega  {{\rm E}_\tau }[log({D_\omega }(s,a))] + {{\rm E}_{{\tau _h}}}[log(1 - {D_\omega }(s,a))]
\end{equation}
The gradient is given by
\begin{equation}\label{5}
{\nabla_\omega }L = {{\rm E}_\tau }[{\nabla_\omega }log({D_\omega}(s,a))] + {{\rm E}_{{\tau _h}}}[{\nabla_\omega }log(1 - {D_\omega }(s,a))]
\end{equation}

\renewcommand{\algorithmicrequire}{ \textbf{Require:}} %Use Input in the format of Algorithm
\renewcommand{\algorithmicensure}{ \textbf{Ensure:}} %UseOutput in the format of Algorithm
\begin{algorithm}
\caption{Hindsight generative adversarial imitation learning}
\begin{algorithmic}[1]
\REQUIRE { Policy(generator) $G_\theta$, discriminator $D_\omega$}
\STATE {Initialize $G_\theta$, $D_\omega$ with random weights $\theta_0$, $\omega_0$}
\STATE {Run $G_{\theta_{0}}$ generating rolled-out trajectories $\tau_0$}
\STATE{Synthesize expert-like demonstration data (Algorithm 1)  $\tau_{h0} \leftarrow Hindsight(\tau_0)$}
\STATE{Pre-train $G_\theta$ using MLE on  $\tau_{h0}$}
\STATE{Pre-train  $D_\omega$ via minimizing cross entropy between $\tau_{h0}$ and  $\tau_0$}
\REPEAT
{
\FOR{ $g_{steps}$}
\STATE{Run policy $G_\theta$  generating rolled-out trajectories $\tau$}
\STATE{Update policy parameter $\theta$ :\\
 $\theta = \theta - \alpha \nabla_{\theta}L$, where $\nabla_{\theta}L$ is shown in equation(3) }
\ENDFOR
\FOR{$d_{steps}$}
\STATE{Use current $G_\theta$ to generate rolled-out trajectories $\tau$}
\STATE{$\tau_{h} \leftarrow Hindsight(\tau)$}
\STATE{Update discriminator parameter $\omega$ :\\
 $\omega = \omega + \beta \nabla_{\omega}L$, where $\nabla_{\omega}L$ is shown in equation(5)}
\ENDFOR
}
\UNTIL{HGAIL converges}
\end{algorithmic}
\end{algorithm}

The fully detailed HGAIL algorithm is shown in Algorithm 2. At the beginning of the training, we generate trajectories ${\tau_0}$ using policy with random weights. Expert-like demonstration data ${\tau_{ho}}$ is synthesized from ${\tau_0}$ with hindsight transformation technique, as is shown in Algorithm 1.  ${\tau_0}$ is regarded as the negative samples and ${\tau_{h0}}$ is considered as positive samples. We use the maximum likelihood estimation (MLE) method to pre-train policy ${G_\theta }$ on ${\tau_{h0}}$. We also pre-train discriminator ${D_\omega}$ via minimizing cross entropy between ${\tau_0}$ and ${\tau_{h0}}$. We found the pre-train procedure is beneficial for the policy training. After the pre-training procedure,
optimizations over policy and discriminator are performed by alternating between policy gradient optimization steps to decrease (2) with respect to policy parameter $\theta$ and gradient step to increase (4) with respect to the discriminator parameter $\omega$. Finally, the policy and the discriminator are both converged.

%%%%%%%%%%%%%%%%%%%%%%%%%%%%%%%%%%%%%%%%%%%%%%%%%%%%%%%%%%%%%%%%%%%%%%%%%%%%%%%%
\section{EXPERIMENTS AND RESULTS}
In this section, our goal is to test whether the policies learned via our proposed HGAIL method works well without external demonstrations provided. In addition, ablation studies are conducted to show the influence of different mechanisms and hyper-parameters on the policy learning. Finally, experiments are carried on to test whether the final learned polies could be directly transfer to real-word physical system.
\subsection{Policy Learning}
To test the feasibility of the proposed HGAIL method, experiments are carried out on two common robot¡¯s tasks: reaching target position and grasping target object \cite{c45} (as is shown in figure 2) in gym \cite{c46} environment. In order to make these two tasks more challenging, we pretend that only binary sparse reward is available in these two tasks. For reaching task, the reward is -1 for most states, and is 0 only when robot gripper reaching the target position. Similarly, for grasping target object task, the reward is -1 for most states, and is 0 only when robot gripper succeeds in grasping the target object.

We compare our proposed HGAIL algorithm  against the following methods: (1) GAIL\cite{c9} with demonstrations available, denoted as GAIL-demo, (2) PPO\cite{c44}, the state-of-the-art of policy gradient method, (3) GASIL\cite{c43}, (4) HGAIL without hindsight transformation technique, denoted as HGAIL-no.

\begin{figure}[h]
\setlength{\abovecaptionskip}{0pt}
\setlength{\belowcaptionskip}{0pt}
	\centering
	\subfigure{% \subfigure[]
		%\label{matchproblem-a}
		\includegraphics[width=3cm, height=3.15cm]{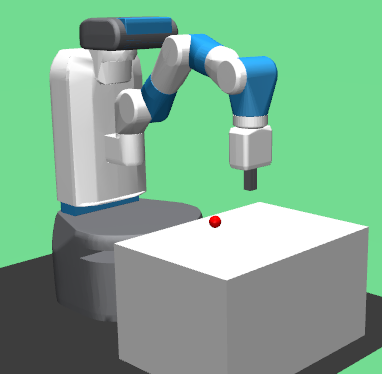}}
	\subfigure{
		%\label{matchproblem-b}
		\includegraphics[width=3cm, height=3.15cm]{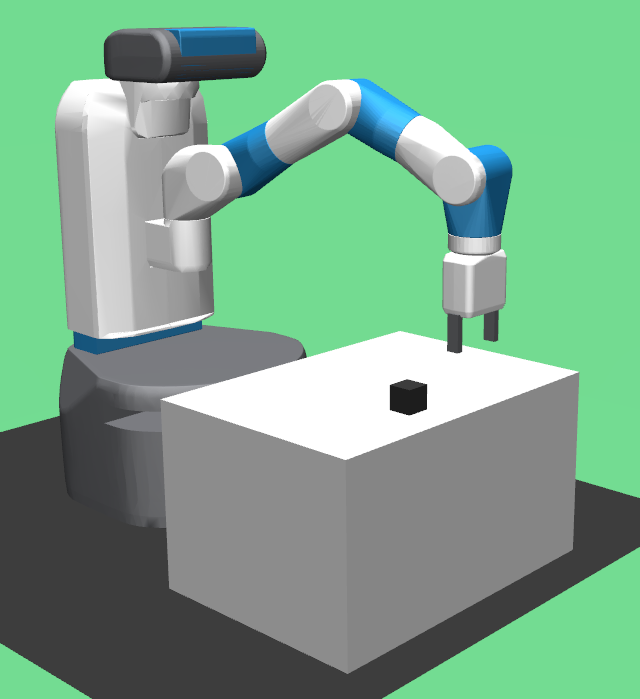}}
	\caption{Two tasks implemented on Fetch robot in gym. Fetch robot owns seven degrees of freedom. In our experiments, the robot takes input as four dimensional action vector, the first three elements of action vector moving the end-effector (gripper) to three orthogonal directions, and the fourth one is controlling the gripper to be closed or open. \textbf{Left}: Reaching task. The red point denotes the target position within the robot workspace. The fourth element of action vector is set to be fixed. \textbf{Right}: Grasping object task. The black cube is the target object to be picked. Best viewed in color}
	\label{fig2}
\end{figure}

The performance of learned policies are measured in terms of two metrics: distance error and success rate. Distance error is measured by the distance between target position  ${p_d}$ and gripper position ${p_g}$ at the end of each episode.
\begin{equation}\label{6}
d = {\left\| {{p_g} - {p_d}} \right\|_1}
\end{equation}
Success rate specifies the ratio of times successfully reaching target positions within allowed error $\delta$ to all times consumed, , as is shown in equation (5) (for reaching task) or grasping the desiered objects (for grasping task) to all times consumed
\begin{equation}\label{7}
{S_{rate}} = \frac{{\sum\limits_i^N {1({d_i} \le \delta )} }}{N}
\end{equation}	
where ${\rm{1}}( \cdot )$ is the indicator function taking true as input and giving 1 as output, and taking false as input and outputting 0.

Implementation details are available in Appendix VI-A. Learning curves on the performance of policies learned with HAGIL compared to above mentioned methods for robot reaching task and grasping tasks are shown in Figure 3 and Table 1 summarizes the performance of the final learned policies.

\begin{figure}[h]
\setlength{\abovecaptionskip}{0.cm}
\setlength{\belowcaptionskip}{0.cm}
	\centering
	\subfigure{% \subfigure[]
		%\label{matchproblem-a}
		\includegraphics[width=4.2cm, height= 3.43cm]{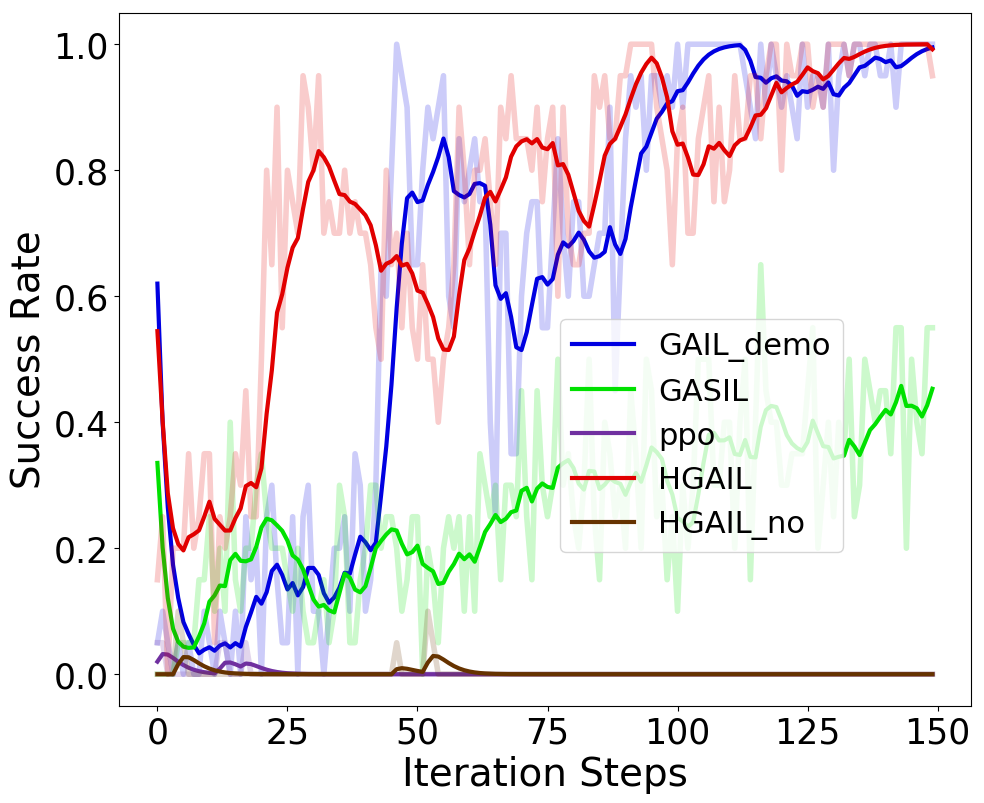}} %width=4.1cm, height= 3.35cm
	\subfigure{
		%\label{matchproblem-b}
		\includegraphics[width=4.2cm, height= 3.43cm]{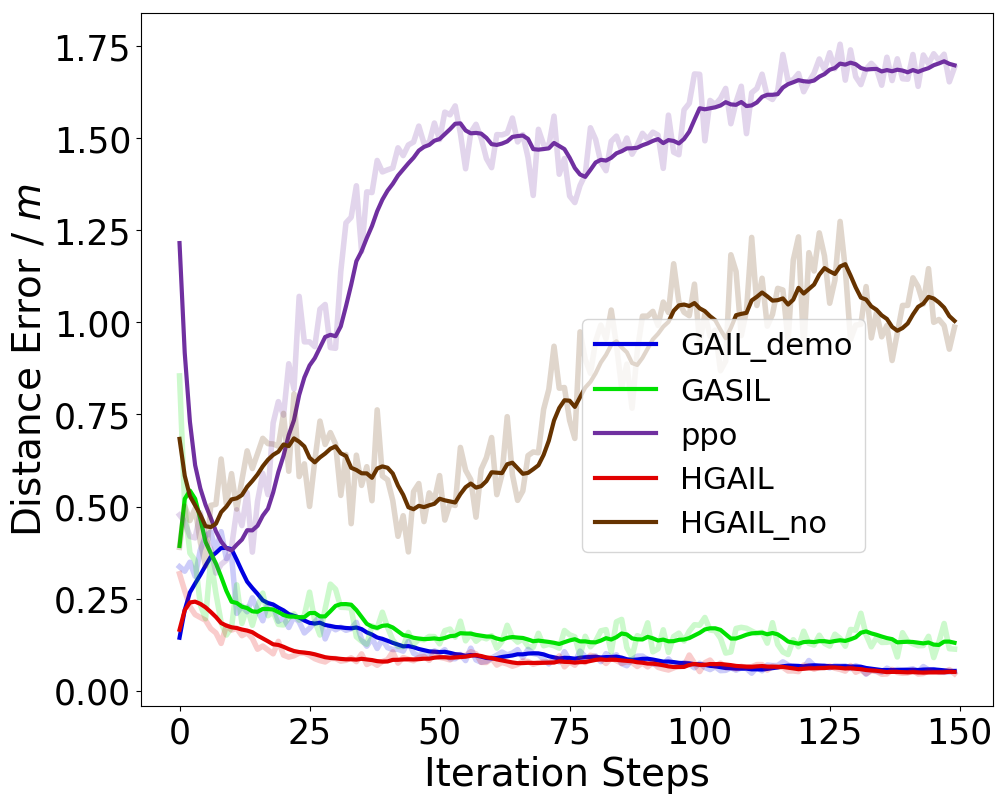}}
\subfigure{% \subfigure[]
		%\label{matchproblem-a}
		\includegraphics[width=4.2cm, height= 3.43cm]{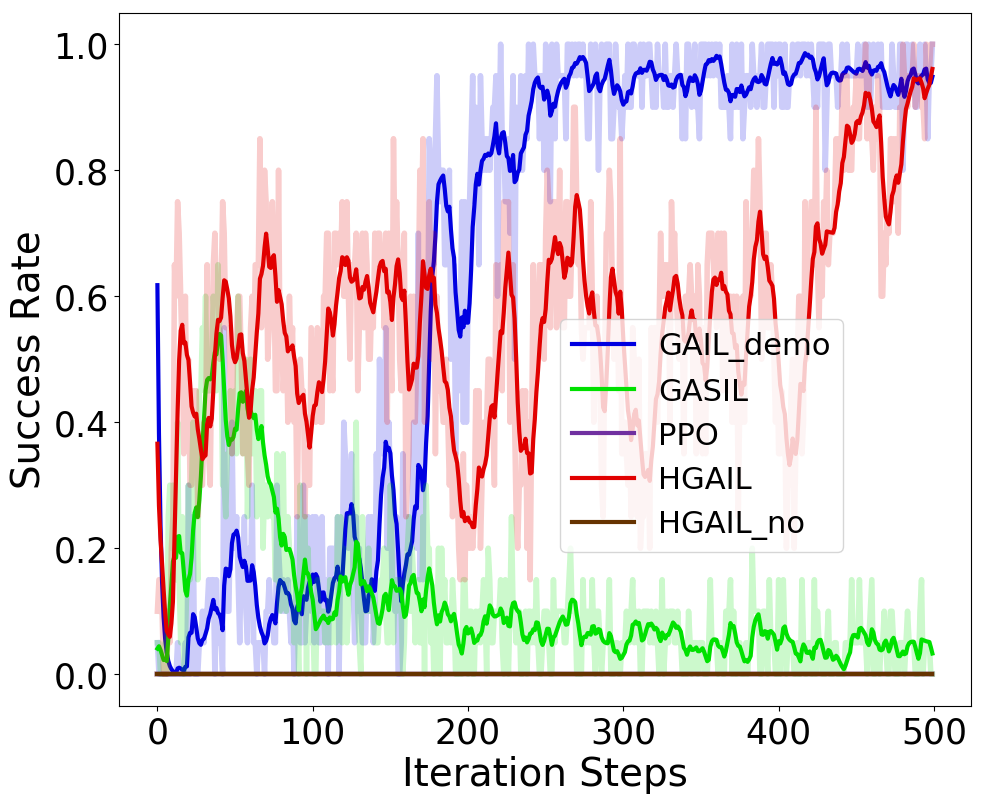}} %width=4.1cm, height= 3.35cm
	\subfigure{
		%\label{matchproblem-b}
		\includegraphics[width=4.2cm, height= 3.43cm]{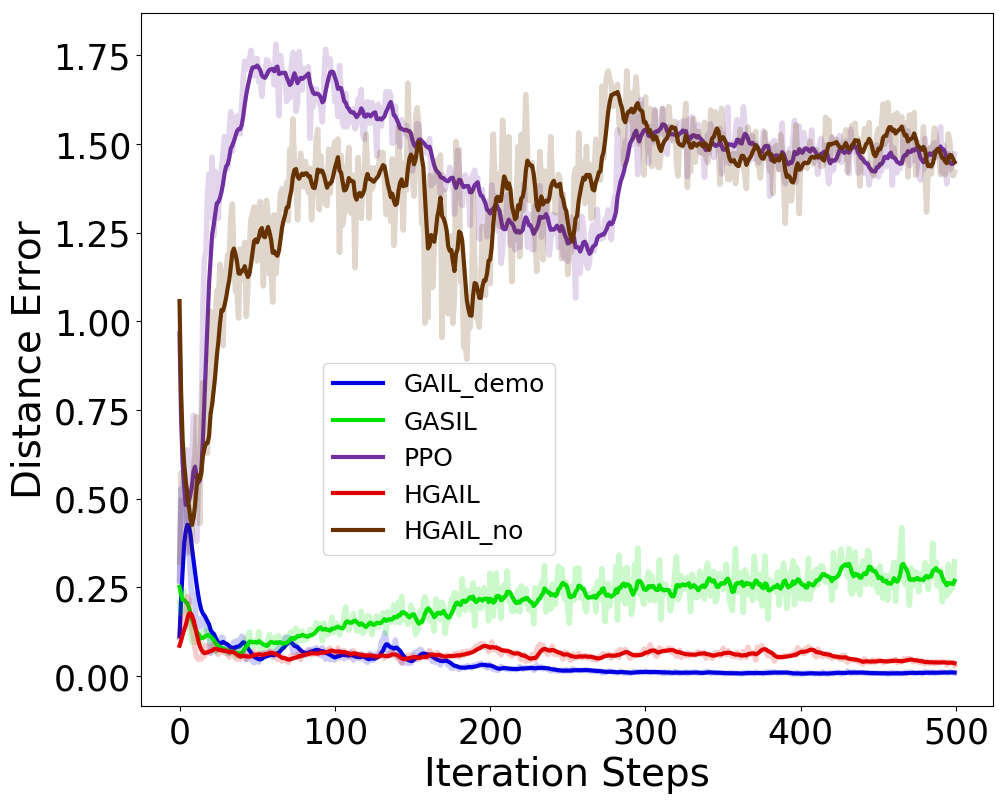}}
	\caption{
 Learning curves on performance of policies learned with HGAIL, GAIL with demonstration data provided (GAIL-demo), PPO, GASIL, and HGAIL without hindsight transformation (HGAIL-no). All the policies use the same network architecture and the same hyper parameters. The first row shows the success rates (left) and distance errors (right) of policies learned for reaching task respectively. The second row shows  the success rates (left) and distance errors (right) of policies learned for grasping object task. Compared to the other algorithms, HGAIL shows the promising performance with no demonstration provided. Best viewed in color. }
	\label{fig3}
\end{figure}

Compared to GAIL with demonstration available, our HGAIL algorithms shows comparable performance in term of success rates and final distance errors for both reaching task and grasping object task. However, the policies trained with our method for grasping task show slower convergence speed. To some extent, without demonstration data, the performance of policies trained with HGAIL are also promising. Compared to PPO, our algorithm shows much better performance. We can draw a conclusion that, although the component of policy in our algorithm is optimized via PPO, out of our HGAIL learning framework, PPO can't train successful policies alone for tasks with binary sparse reward. Policies trained with GASIL show slower optimization speed and poorer polices performance in these two tasks. In comparison with HGAIL-no, HGAIL exhibits better performance, which indicates that hindsight transformation is crucial ingredient in our proposed HGAIL algorithm when demonstrations are not available. The results prove that HGAIL can work well without demonstration data available, and learn successful policies. We can also see that, as our algorithm essentially endows curriculum learning mechanism, at the beginning period of policy training, HGAIL shows the faster optimization than GAIL algorithm with demonstrations available.

%
%We are curious to know wherther our HGAIL algorithm can show comparable performance to gail with deomnstration data provided.
%
%As our final learned policy  is trained via PPO in our proposed framework,
%
%The purpose of this comparison is to exclude the influence of  PPO algorithm alone on  performance of the learned policy under our HGAIL learning framework.
%
%GASIL is a recent new algorithm, improving existing state-of-the-art baselines across many contol tasks. Comparasion experiments is for testing whether our algorithm outperforms GASIL in high continuous action space and state space, for example in reaching and grapsping tasks.

\begin{table}
\setlength{\abovecaptionskip}{-2pt}
\setlength{\belowcaptionskip}{-5pt}
\caption{ Performance of policies trained with different algorithms}
\label{table7}
\begin{center}
\setlength{\tabcolsep}{0.1mm}{
\begin{tabular}{|c|c|c|c|c|}
\hline
\multirow{2}{*}{Method} & \multicolumn{2}{|c|}{Reaching task}  & \multicolumn{2}{|c|}{Grasping object task} \\
\cline{2-5}
& success rate & distance error $(m)$ & success rate & distance error $(m)$\\

\hline
GAIL-demo         & $0.98\pm0.02$    &${0.03\pm0.01 }$ &$0.95\pm0.02$    &${0.02\pm0.003 }$ \\
\hline
GASIL            & $0.42\pm0.08$    &${0.20\pm0.07}$   & $0.03\pm0.03$    &${0.25\pm0.06}$\\
\hline
PP0              & $0$               &${1.70\pm0.12}$   & $0$              &${1.53\pm0.08}$ \\
\hline
HGAIL(ours)       & $0.98\pm0.02$    &${1.05\pm0.03}$   & $0.95\pm0.03$    &${0.03\pm0.002}$\\
\hline
HGAIL-no          & $0.01\pm0.01$    &${0.62\pm0.27}$   & $0.02\pm0.02$    &${1.49\pm0.04}$\\
\hline

\end{tabular}}
\end{center}
\end{table}

%\begin{table}[h]
%\caption{ Comparing policies performance with different training methods}
%\label{table1}
%\begin{center}
%\begin{tabular}{|c|c|c|}
%\hline
%Method & Success rates & Distance errors ($m$) \\
%\hline
%GAIL-demos\cite{c9}             & $0.98\pm0.02$           &${0.03\pm0.004 }$ \\
%\hline
%GASIL\cite{c43}                    & $0.62\pm0.07$           &${0.16\pm0.07}$ \\
%\hline
%PP0\cite{c44}                     & $ 0.38\pm0.14$           &${0.18\pm0.12}$ \\
%\hline
%HGAIL(ours)      & $0.98\pm0.02$           &${0.03\pm0.01}$ \\
%\hline
%HGAIL-no      & $0.26\pm0.05$           &${0.62\pm0.27}$ \\
%
%\hline
%\end{tabular}
%\end{center}
%\end{table}

%
%
%\begin{table}[h]
%\caption{ Comparing policies performance with different training methods}
%\label{table1}
%\begin{center}
%\begin{tabular}{|c|c|c|c|c|}
%\hline
%Method & Success rates & Distance errors ($m$) & Success rates & Distance errors ($m$)\\
%\hline
%GAIL-demos\cite{c9}             & $0.98\pm0.02$           &${0.03\pm0.004 }$  & $0.98\pm0.02$           &${0.03\pm0.004 }$\\
%\hline
%GASIL\cite{c43}                    & $0.62\pm0.07$           &${0.16\pm0.07}$  & $0.98\pm0.02$           &${0.03\pm0.004 }$ \\
%\hline
%PP0\cite{c44}                     & $ 0.38\pm0.14$           &${0.18\pm0.12}$   & $0.98\pm0.02$           &${0.03\pm0.004 }$\\
%\hline
%HGAIL(ours)      & $0.98\pm0.02$           &${0.03\pm0.01}$   & $0.98\pm0.02$           &${0.03\pm0.004 }$\\
%\hline
%HGAIL-no      & $0.26\pm0.05$           &${0.62\pm0.27}$   & $0.98\pm0.02$           &${0.03\pm0.004 }$\\
%
%\hline
%\end{tabular}
%\end{center}
%\end{table}

\subsection{Ablation Studies}
  In our experiments, ablation studies on reaching task and grasping target object task show the similar conclusions. Consequently, in order to make the content of the paper more concise and compact, by default, we mainly show the results of experiments on reaching task in this section.
\subsubsection{Curriculum Learning or Not}
In HGAIL learning framework, hindsight transformed data (expert-like data) is converted from various levels of data rolled out by different-levels' generator in the procedure of adversarial learning. To some extent, our HGAIL learning paradigm essentially endows training agent policy with curriculum learning mechanism.

In order to show whether this curriculum learning mechanism is crucial for  policy training,  experiments on policies trained without curriculum learning mechanism is conducted. Concretely, in the ablation experiments, hindsight transformed data (expert-like data) are transformed only from rolled-out trajectories ${\tau_0}$ produced by the policy $G_{\theta_{0}}$ at the beginning of training period. Learning curves on success rates and distance errors are shown in Figure 4 and Table 2 summaries the performance of the final trained policies. As illustrated, policy trained with curriculum learning mechanism shows the better performance with respect to both success rate and distance error, and the learning process is more stable.

\begin{figure}[h]
\setlength{\abovecaptionskip}{0.cm}
\setlength{\belowcaptionskip}{0.cm}
	\centering
	\subfigure{% \subfigure[]
		%\label{matchproblem-a}
		\includegraphics[width=4.2cm, height= 3.43cm]{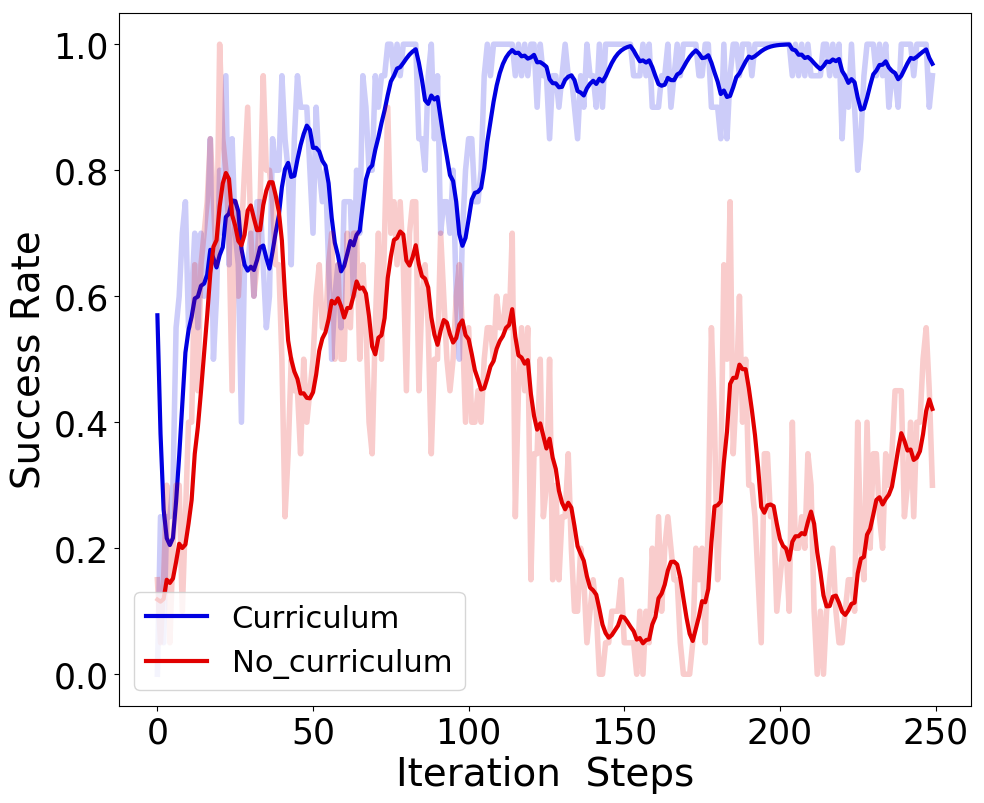}}
	\subfigure{
		%\label{matchproblem-b}
		\includegraphics[width=4.2cm, height= 3.43cm]{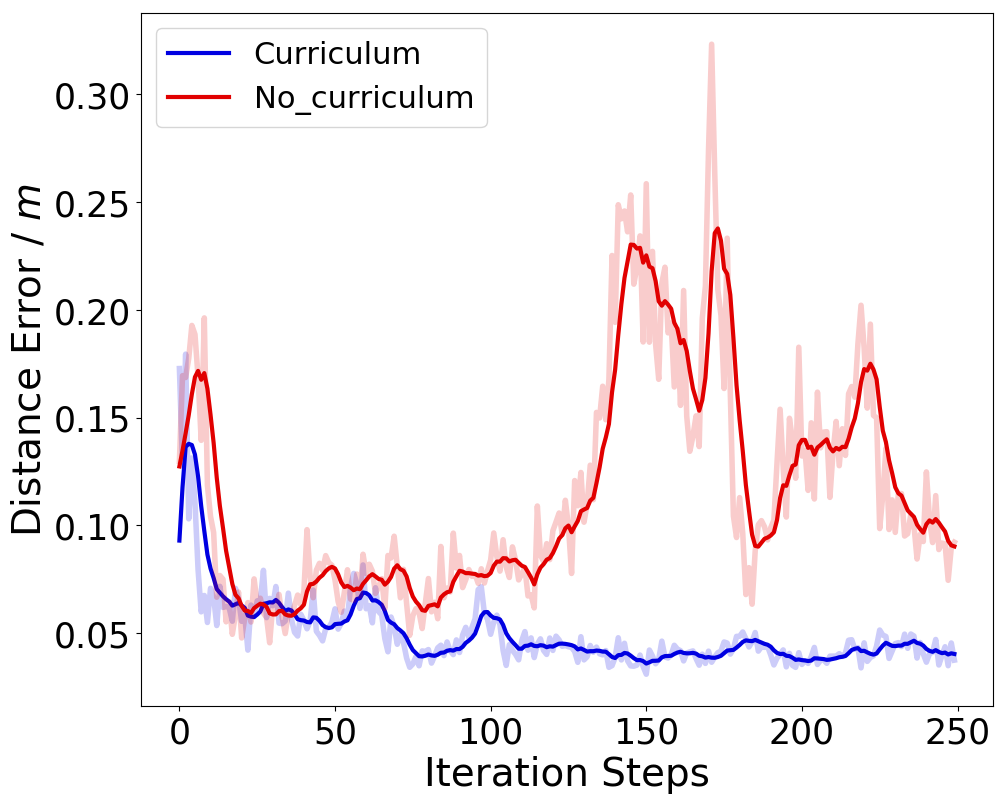}}
	\caption{Learning curves on policy performance with respect to employing curriculum learning mechanism or not vs iteration steps. \textbf{Left}: success rate. \textbf{Right}: distance errors. As is shown, policies trained with curriculum shows the better performance. At the same time, the training process with curriculum mechanism is more stable. Best viewed in color.}
	\label{fig4}
\end{figure}

\begin{table}[h]
\caption{ Performance of policies trained with curriculum learning mechanism or not}
\label{table2}
\begin{center}
\begin{tabular}{|c|c|c|}
\hline
Method & Success rates & Distance errors ($m$) \\
\hline
Curriculum learning             & $0.98\pm0.02$           &${0.03\pm0.01}$ \\
\hline
No curriculum learning     & $0.49\pm0.17$           &${0.16\pm0.05}$ \\
\hline
\end{tabular}
\end{center}
\end{table}

\subsubsection{Formation of Hindsight Transformation}

Inspired from HER \cite{c14}, we propose two different strategies for hindsight transformation called \emph{final hindsight transformation} and \emph{future hindsight transformation} respectively. \emph{Final hindsight transformation} replaces the goal of each state with the position of the final reached state in its own episode. However, \emph{Future hindsight transformation} is randomly changing target position of each state with the position of state observed after it, as shown in Algorithm 1.

Learning curves in terms of two different hindsight transformation are shown in Figure 5 and Table 3 summarizes the final policies performance. Final hindsight transformation can¡¯t work well and the policy learned with it divergent gradually in the training procedure.

\begin{figure}[h]
\setlength{\abovecaptionskip}{0.cm}
\setlength{\belowcaptionskip}{0.cm}
	\centering
	\subfigure{% \subfigure[]
		%\label{matchproblem-a}
		\includegraphics[width=4.2cm, height= 3.43cm]{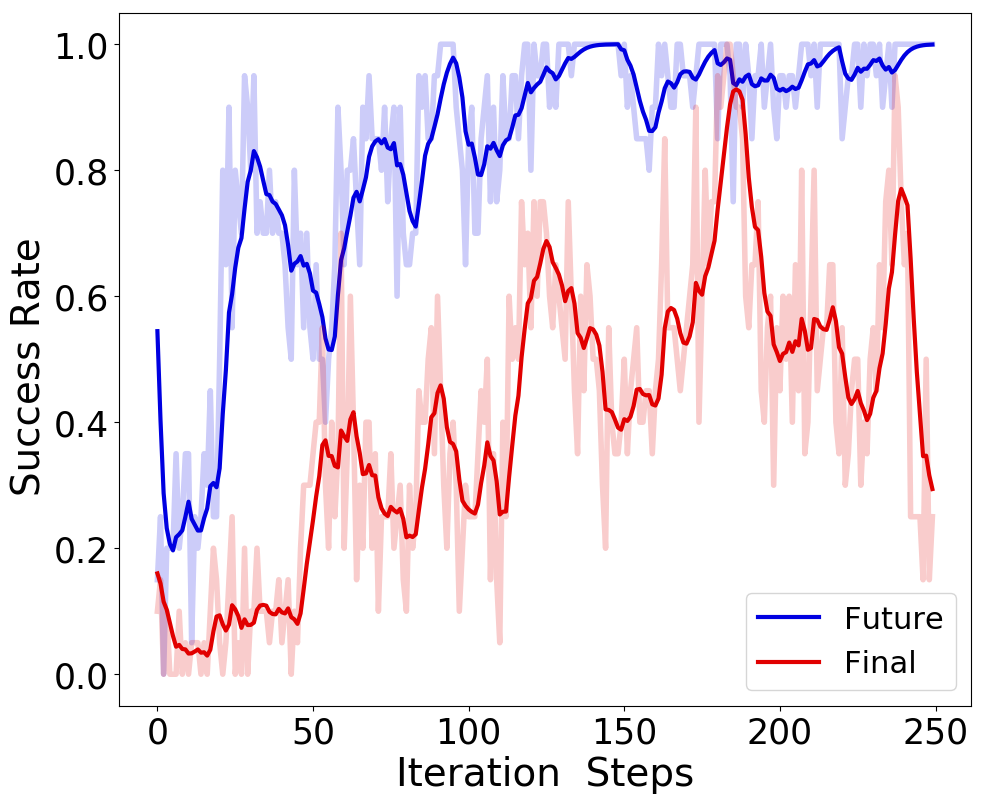}}
	\subfigure{
		%\label{matchproblem-b}
		\includegraphics[width=4.2cm, height= 3.43cm]{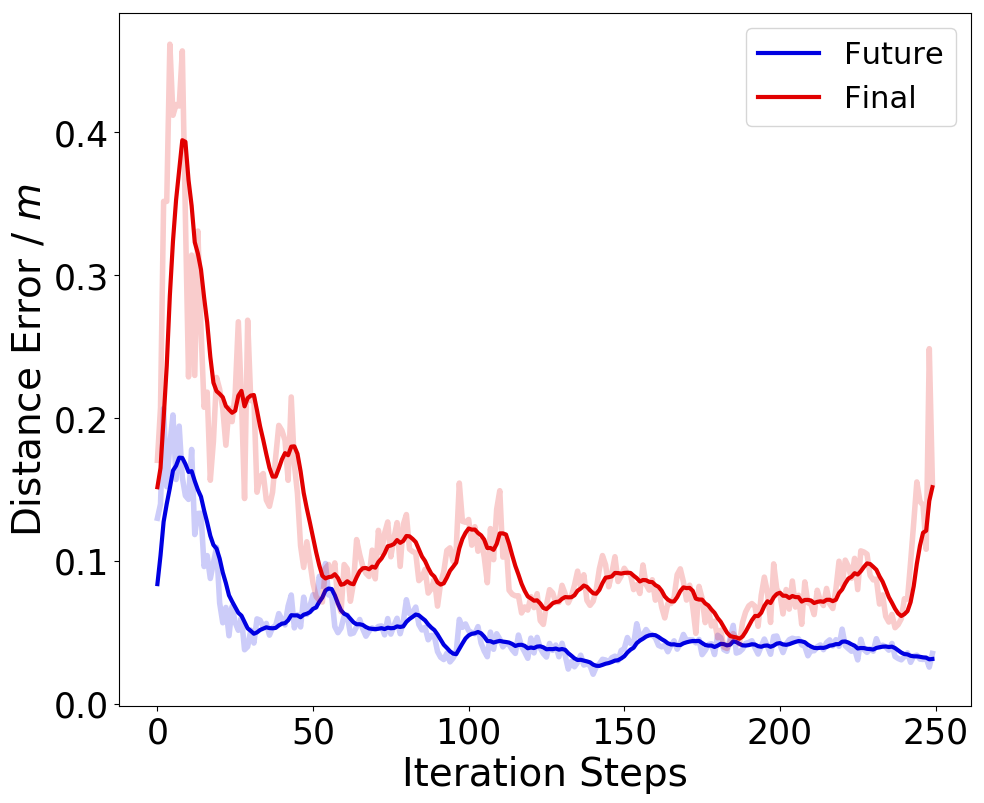}}
	\caption{ Learning curves on policy performance based on future hindsight transformation (Future) and  final  hindsight transformation (Final) vs iteration steps. \textbf{Left}: success rates. \textbf{Right}: distance errors. Results show that policy trained with future hindsight transformation exhibits better performance. The training procedure is more stable with future hindsight transformation. Best viewed in color.}
	\label{fig5}		
\end{figure}

\begin{table}[h]
\setlength{\abovecaptionskip}{0pt}
\setlength{\belowcaptionskip}{0pt}
\caption{ Performance of  policies trained with different hindsight transformation}
\label{table3}
\begin{center}
\begin{tabular}{|c|c|c|}
\hline
Hindsight transformation & Success rates & Distance errors ($m$) \\
\hline
Future            & $0.99\pm 0.01$           &${0.04\pm 0.01}$ \\
\hline
Final   & $0.31\pm 0.17$           &${0.18\pm 0.05}$ \\
\hline
\end{tabular}
\end{center}
\end{table}
\subsubsection{Hindsight Transformation Probability}

So far, the performance of all the learned policies were trained with hindsight transformation probability $p_{ht} = 1$ . We also interested in the effect of  $p_{ht}$  value on the performance of the final learned policy. Experiments are carried out with $p_{ht}$   being 0.2, 0.4, 0.6, 0.8 and 1. Learning curves are shown in Figure 6. The results illustrates that converting each state into hindsight transformation with probability 1 performs best, which is different from HER \cite{c14}. The bigger value of the hindsight transformation probability is, the better performance the final learned policy demonstrates.

\begin{figure}[h]
\setlength{\abovecaptionskip}{0.cm}
\setlength{\belowcaptionskip}{0.cm}
	\centering
	\subfigure{% \subfigure[]
		%\label{matchproblem-a}
		\includegraphics[width=4.2cm, height= 3.43cm]{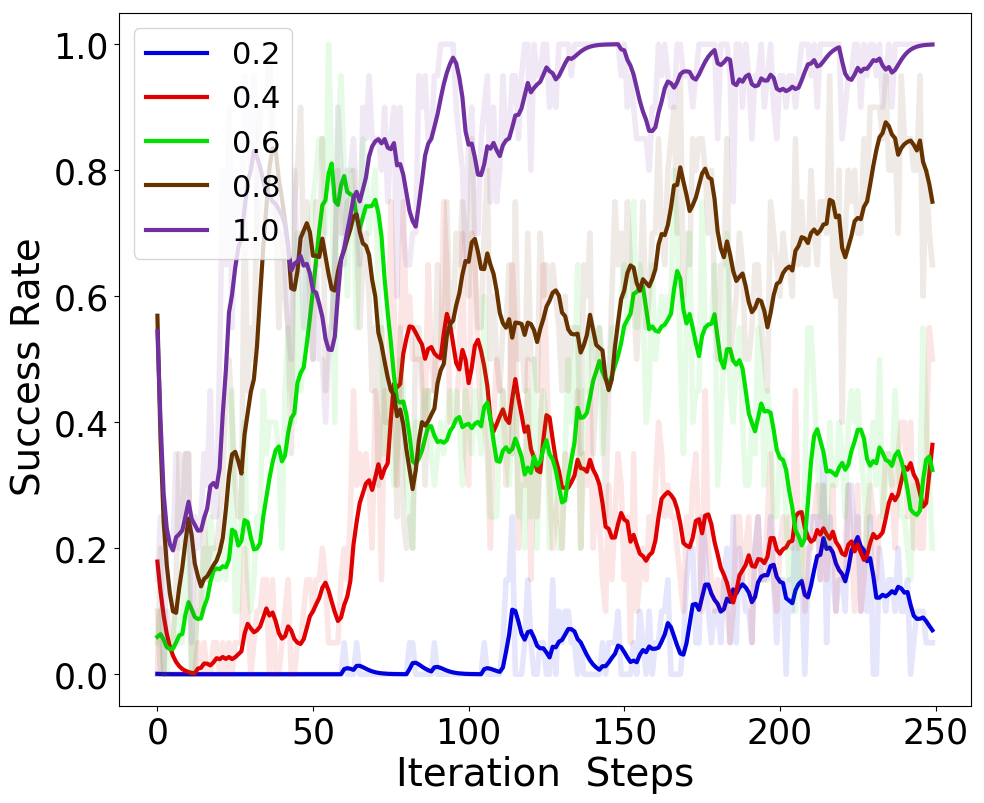}}
	\subfigure{
		%\label{matchproblem-b}
		\includegraphics[width=4.2cm, height= 3.43cm]{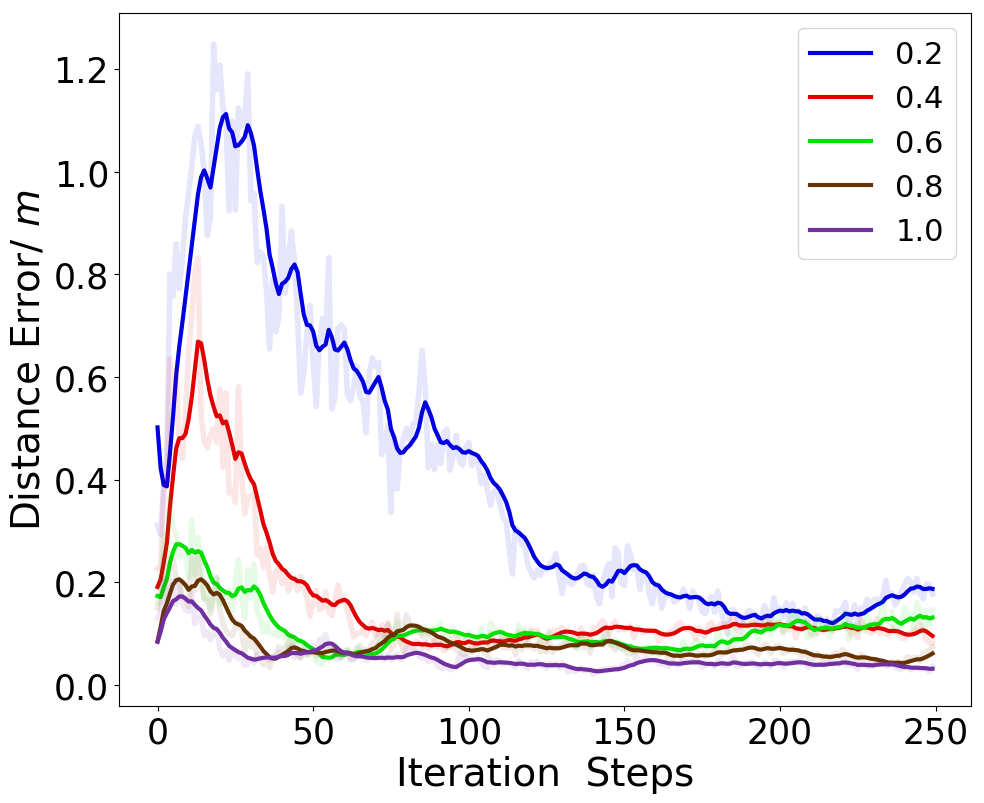}}
	\caption{Learning curves on the performance of  policies  learned with $p_{ht}$ being 0.2, 0.4, 0.6, 0.8, and 1 vs iteration steps.  \textbf{Left}: success rates. \textbf{Right}: distance errors. Policies learned with hindsight transformation probability being 1 shows the best performance in terms of success rate and distance error. The bigger the hindsight transformation probability is, the better performance the trained policy demonstrates. Best viewed in color. }
	\label{fig6}		
\end{figure}

\subsubsection{Reward Formation}

Different reward function for policy learning in HGAIL learning framework is experimentally analyzed. In the GAIL learning framework, different reward formations for ${r_\omega }({s_t},{a_t})$ has been applied [9][24]. We compare four common reward functions written as
${r_1}({s_t},{a_t}) =  - log{\rm{(}}1 - sig{\rm{(}}dis{\rm{(}}{s_t},{a_t}{\rm{))}}$,
${r_2}({s_t},{a_t}) = log{\rm{(clip(}}dis{\rm{(}}{s_t},{a_t}{\rm{), 0,1))}}$, ${r_3}({s_t},{a_t}) = dis{\rm{(}}{s_t},{a_t}{\rm{)}}$, $ {r_4}({s_t},{a_t}) = log{\rm{(}}sig{\rm{(}}dis{\rm{(}}{s_t},{a_t}{\rm{))}}-log{\rm{(}}1 - sig{\rm{(}}dis{\rm{(}}{s_t},{a_t}{\rm{))}}$,
where $sig$ denotes sigmoid function,  $dis{\rm{(}}{s_t},{a_t}{\rm{)}}$ is the output of discriminator taking state ${s_t}$  and action  ${a_t}$ pairs as input, and  ${\rm{clip(}}dis{\rm{(}}{s_t},{a_t}{\rm{), 0,1)}}$is clipping $dis{\rm{(}}{s_t},{a_t}{\rm{)}}$ to $0\sim 1$. The results are illustrated in Figure 7. The policies learned from reward ${r_1}({s_t},{a_t})$ converged fastest compared to other three reward functions. ${r_1}({s_t},{a_t})$, ${r_3}({s_t},{a_t})$, and ${r_4}({s_t},{a_t})$ guides the final learned policies exhibits similar better performance in terms of distance error in contrast to ${r_2}({s_t},{a_t})$ . The policies learned from ${r_1}({s_t},{a_t})$ show the best performance with respect to not only in iteration steps consumed for policy training, but also in higher success rates and lower distance errors. As a result, in our work, we choose ${r_1}({s_t},{a_t})$  as our default reward function for policy learning.

\begin{figure}[h]
\setlength{\abovecaptionskip}{0.cm}
\setlength{\belowcaptionskip}{0.cm}
	\centering
	\subfigure{% \subfigure[]
		%\label{matchproblem-a}
		\includegraphics[width=4.2cm, height= 3.43cm]{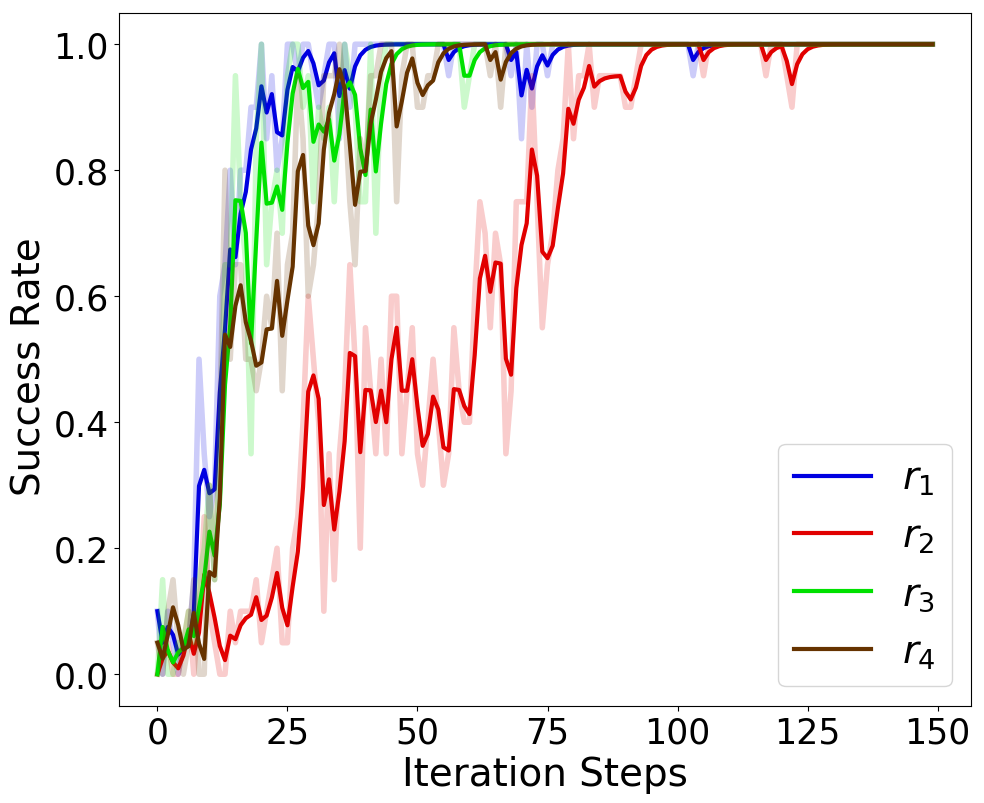}}
	\subfigure{
		%\label{matchproblem-b}
		\includegraphics[width=4.2cm, height= 3.43cm]{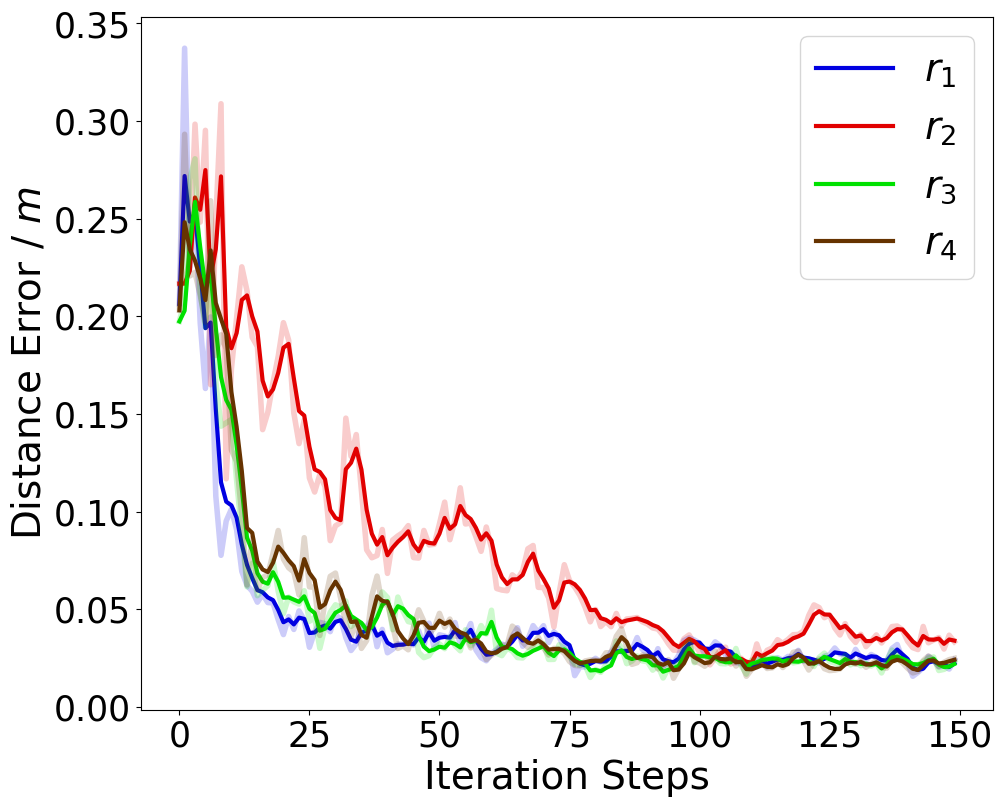}}
	\caption{Learning curves on policy performance with respect to four different reward functions vs iteration steps. \textbf{Left}: success rates. \textbf{Right}: distance errors. As is shown, policies learned from ${r_1}({s_t},{a_t})$ show the best performance not only in iteration steps consumed for policy training converged, but also in higher success rates and lower distance error. Best viewed in color. }
	\label{fig7}
\end{figure}

\subsection{Sim to Real Policy Transfer}
To validate the feasibility of the policy trained with our algorithm deployed in real-world physical system (no additional training). Experiment are conducted on real-world UR5 robot (the only robot arm available in our lab). The detail implementation of experiments is shown in  Appendix VI-B. As is shown in Figure 8,  the position of red ball is the target position for reaching task, and The pink cube is the target object to be grasped for grasping object task.
Frames of UR5 robot employing learned policy in reaching target position and grasping target object are pictured respectively, as is shown in Figure 8. Success rates and distance errors are summarized in Table 4. Results show that policy learned with HGAIL can successfully transfer from simulated environment to real-word scenarios, and the performance in real-world scenarios is consistent with simulated environment without additional training.
\begin{table}[h]
\setlength{\abovecaptionskip}{0pt}
\setlength{\belowcaptionskip}{0pt}
\caption{ Performance of  policies employed in real-word scenarios}
\label{table4}
\begin{center}
\begin{tabular}{|c|c|c|}
\hline
Task & Success rates & Distance errors ($m$) \\
\hline
Reaching             & $0.95\pm 0.03$           &${0.01\pm 0.01}$ \\
\hline
Picking     & $0.93\pm 0.04$           &${0.02\pm 0.01}$ \\
\hline
\end{tabular}
\end{center}
\end{table}

\begin{figure}[h]
\centering
\includegraphics[scale=0.95]{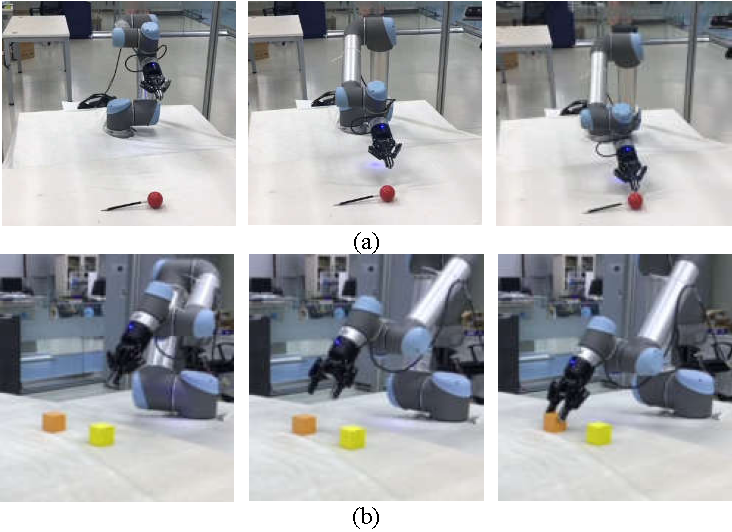}
\caption{ Frames of learned policy employed on real-world UR5 robot for reaching task (a) and grasping target object (b). We using the red ball define the target position for reaching task. The pink cube is the target object to be grasped for grasping object task. Robot succeeded in reaching target position and grasping the target object with high accuracies.}
\label{fig9}
\end{figure}

%%%%%%%%%%%%%%%%%%%%%%%%%%%%%%%%%%%%%%%%%%%%%%%%%%%%%%%%%%%%%%%%%%%%%%%%%%%%%%%%
\section{CONCLUSION}
We propose HGAIL algorithm which is a new learning paradigm under GAIL learning framework for learning control policy without expert demonstration available. We adopt hindsight transformation mechanism to self- synthesize expert-like demonstration data for adversarial policy learning. Experimental results show that the proposed method efficiently and effectively trains policies. In addition, hindsight transition technique essentially endowing curriculum learning mechanism under our learning framework is critical for policy learning. We also validate the feasibility of the policy trained with our algorithm directly deployed in real-world robot without additional training.

In the future, we want to employ our method in more continuous and discrete environments. A promising line is directly applying our method in training manipulation skills on real-world robot, as the amount of training interaction data is relatively small. Another exciting direction is to combine the HGAIL algorithm with hierarchy to solve more complicated tasks.

\section{Appendix}
\subsection{Implementation Details}
In this section, we provide additional details about the experimental tasks setup and hyper-parameters.
\subsubsection{Generator}
We use two layer tanh neural network with 64 units for the value network and policy network. The policy network take input as a concatenated vector with gripper position, gripper velocity, and target position. The policy network¡¯s out parameterizes the Gaussian policy distribution, where the mean is the output of the policy network and the fixed covariance was set to be 1.
\subsubsection{Discriminator}
We use two-layer tanh neural network with 100 units in each layer for the discriminator $D$.

We set $g_{steps}=16$, $d_{steps}=3$. Learning rate for discriminator is 0.0004, and learning rate for generator is $\lambda=0.001$. Batch size is set to be 64 for discriminator optimization, and 128 for generator optimization. The pre-train steps for generator is 100 and for discriminator is 500. For fair comparison, all experiments were run in a single thread, all of the algorithms( HGAIL, GAIL-demo, GASIL, and HGAIL-no) shares the same network architecture and the same hyper parameters and PPO share these parameters with generator.

It should be mentioned that, if not clearly indicated in the paper, all our parameters about ablation studies are set to the following default values: Hindsight transformation probability $p_{ht} = 1$, reward function is  ${r_{dis}}({s_t},{a_t})={r_1}({s_t},{a_t}) =  - {\rm{log(1}} - sig{\rm{(}}dis{\rm{(}}{s_t},{a_t}{\rm{))}}$.

\subsection{Transfer to real-world robot}
The final learned policy is directly transferred from simulated environment to real-word UR5 robot without additional training. As show in figure 8, we use different object to define the target position. In our working scenario, RGB-D image can be obtained from depth camera installed above the robot. Another trained deep neural network (VGG-16) output object pixel position $(u, v)$. The target object position $p=(x, y, z)$ under robot coordinate systems can be obtained by the following equation
\begin{equation}\label{8}
\left[ {\begin{array}{*{20}{c}}
x\\
y\\
z
\end{array}} \right] = RzM_{in}^{ - 1}\left[ {\begin{array}{*{20}{c}}
u\\
v\\
1
\end{array}} \right] + T
\end{equation}
where $M_{in}$ is camera inner parameter matrix, $z$ is depth value with respect to the pixel position $(u, v)$, and $R$ and $T$ are the rotation matrix and transformation vector from the camera coordinate system to the robot coordinate system respectively.
At time step $t$, the gripper¡¯s position $p_t$, velocity $v_t$, target object position p are contacted into a single vector $[{p_t},{v_t},p]$ fed into the policy network, which is similar to training of fetch arm in simulated environment. The mean of the output of Gaussian policy is send to robot controller, and UR5 robot gripper moved to the next step position $p_{t+1}$. The above procedure is repeated until the ending of the episode.

\addtolength{\textheight}{-12cm}   % This command serves to balance the column lengths

\end{document}